\documentclass{article}

\usepackage[nonatbib, final]{neurips_2023} % final
\usepackage[utf8]{inputenc} % allow utf-8 input
\usepackage[T1]{fontenc}    % use 8-bit T1 fonts
\usepackage{hyperref}       % hyperlinks
\usepackage{url}            % simple URL typesetting
\usepackage{booktabs}       % professional-quality tables
\usepackage{amsfonts}       % blackboard math symbols
\usepackage{nicefrac}       % compact symbols for 1/2, etc.
\usepackage{microtype}      % microtypography
\usepackage{xcolor}         % colors
\usepackage{graphicx}
\usepackage{caption}
\usepackage{subcaption}
\usepackage{enumerate}

\usepackage{biblatex}
\usepackage{wrapfig}

\title{Search Strategies for Self-driving Laboratories with Pending Experiments}

\author{%
  Hao Wen \\
  University of Surrey, UK\\
  Matterhorn Studio\\
  \texttt{hao@matterhorn.studio} \\
  \And
  Jakob Zeitler \\
  Matterhorn Studio\\
  Oxford, United Kingdom \\
  \texttt{jakob@matterhorn.studio} \\
  \And
  Connor Rupnow \\
  \texttt{connor@rupnow.com} \\
}

\begin{document}

\maketitle

\begin{abstract}
Self-driving laboratories (SDLs) consist of multiple stations that perform material synthesis and characterisation tasks. To minimize station downtime and maximize experimental throughput, it is practical to run experiments in asynchronous parallel, in which multiple experiments are being performed at once in different stages. Asynchronous parallelization of experiments, however, introduces delayed feedback (i.e. “pending experiments”), which is known to reduce Bayesian optimiser performance. Here, we build a simulator for a multi-stage SDL and compare optimisation strategies for dealing with delayed feedback and asynchronous parallelized operation. Using data from a real SDL, we build a ground truth Bayesian optimisation simulator from 177 previously run experiments for maximizing the conductivity of functional coatings. We then compare search strategies such as expected improvement, noisy expected improvement, 4-mode exploration and random sampling. We evaluate their performance in terms of amount of delay and problem dimensionality. Our simulation results showcase the trade-off between the asynchronous parallel operation and delayed feedback. 

\end{abstract}

%Claim 1: Parallel operation increases the experimental throughput at the cost of optimization performance
%Claim 2: Increasing the number of dimensions in a BO problem decreases performance
%Claim 3: 

% Instructions: We encourage submissions of short-form papers up to 5 pages in length with unlimited pages for references and supplementary materials. Submissions in this track should clearly explain how the proposed work helps accelerate material discovery in cases where the relation may be ambiguous. Reviewers will be asked to evaluate the thoroughness and quality of technical work described in the submission. (from: https://sites.google.com/view/ai4mat/submissions?authuser=0)

\section{Introduction}
% link to google doc where I am working on the intro: https://docs.google.com/document/d/1T3Iwub-mkTXHxce1y81SYy1eQuxoxZ8TqwgN-pHnjKA/edit?usp=sharing 

\begin{figure}

         \centering
         \includegraphics[width=\textwidth]{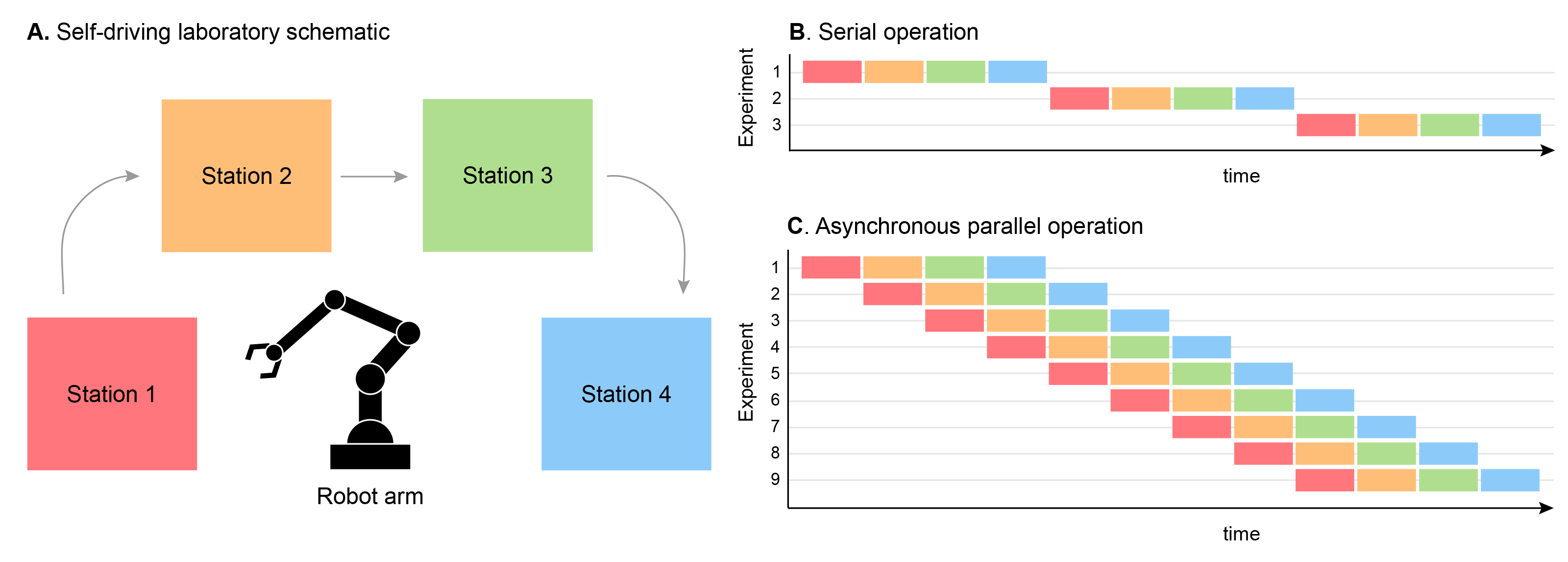}

         \label{fig:setup}
     \hfill

    \caption{ \textbf{(A)} 
Schematic of an SDL with four stages. The robot arm moves experiments from stage to stage. \textbf{(B)} In \textit{serial} operation, the SDL performs each task in order and waits for one experiment to finish before starting the next. \textbf{(C)} In \textit{asynchronous parallel} ("conveyor belt") operation, the SDL begins a new experiment as soon as the first stage is available. In this way, no stage is sitting idle for a significant amount of time and the experimental throughput is significantly increased. In the example above, 9 experiments are completed in asynchronous parallel operation in the same time it took to complete 3 in serial operation. Asynchronous parallel operation, however, causes a delay in the return of data as experiment N+3 must begin before experiment N is complete.}
    \label{fig:conveyor_belt}
\end{figure}

Self-driving laboratories (SDLs) utilize automation and artificial intelligence (AI) to accelerate the discovery of new mission critical materials  such as perovskites \cite{KIRMAN2020938}, catalysts \cite{burger2020mobile},  nanoparticles \cite{autonomous_quantum_dot}, and more \cite{mech_design} \cite{flowsynth} \cite{thinfilms} \cite{snapp2023autonomous}. SDLs consist of multiple stages that, step by step, perform different material synthesis and characterisation tasks (Figure \ref{fig:conveyor_belt}A)\cite{rupnow2023self}. 

As the SDL performs an experiment, the material moves between task stations much like a conveyor belt in a production line. The simplest way to make materials would be to run experiments one at a time in serial operation (Figure \ref{fig:conveyor_belt}B). To minimize station downtime and maximize experimental throughput, it is practical to have all stations running continuously.  This approach is known as \textit{asynchronous parallel}, in which multiple experiments are being performed at once in different stages (Figure \ref{fig:conveyor_belt}C). This increases experimental throughput by allowing the SDL to perform multiple experiments at once, rather than one at a time. 

Asynchronous parallelization of experiments, however, introduces delayed feedback in which new experiments must be chosen while previous experiments are running. The optimisation algorithm must use incomplete information to choose new experimental conditions. Delayed feedback is known to reduce Bayesian optimiser performance \cite{pmlr-v97-alvi19a}. Additionally, the number of processing variables (i.e. dimensions) in an experiment can increase the difficulty of finding an optimal material. To the best of our knowledge, the literature has no empirical results or discussion on the impact of pending experiments for SDLs.

In this work, we fill this gap by building a simulator for a multi-stage SDL and comparing optimisation strategies for handling delayed feedback from asynchronous parallelized operation. We compare the performance of four optimisation strategies as well as compare their performance when subject to different amounts of delay and different number of dimensions on various test functions. Our simulation results showcase the trade-off between the efficiency gains of asynchronous parallel operation and performance loss due to delayed feedback. 

\section{Setup and Experiments}

We built a simulator for a multi-stage SDL and compare optimisation strategies for dealing with delayed feedback and asynchronous parallelized operation.  The simulator is based on the basic Bayesian optimisation example in BoTorch \cite{botorch_tutorial}.  The simulated experiments are performed by testing a set of x-values on a synthetic experiment defined by a function. Noise is artificially added to the result to mimic experimental noise. To mimic the effects of using multiple stages in asynchronous parallel, the results are withheld from being returned until a predefined number of experiments have been completed. A delay of N would represent the asynchronous parallel operation of an SDL with N+1 stages. It should be noted that SDLs can typically operate with any number of experiments running in parallel, capped by the number of stages it has. With the simulator, we run synthetic experiments on three different test functions that are supposed to represent different material's behaviour: 
\begin{enumerate}[(1)]
    \item \textbf{Ackley test function:} an optimisation test function resembling a "needle-in-a-haystack" problem \cite{siemenn2023fast} whose dimensions can be scaled up and down (Appendix Figure \ref{fig:test_functions}A and equation \ref{eq:Ackley}).
    \item \textbf{Levy test function:} a reasonably challenging common optimisation test function whose dimensions can be scaled up and down (Appendix Figure \ref{fig:test_functions}B and equation \ref{eq:Levy}).
    \item \textbf{SDL test function: }Using real SDL data from \cite{rupnow2023self}, we build a ground truth Gaussian Process (GP)\cite{RasmussenW06} model from 177 previously run SDL experiments for maximizing the conductivity of functional coatings. 
\end{enumerate}

We choose the Ackley and Levy test functions as they are established benchmarks in the Bayesian Optimisation literature \cite{buathong2023bayesian,10.3389/fams.2022.1076296, wang2020parallel}, and also allow evaluation across increasing numbers of dimensions. The SDL test function is designed to mimic the natural behaviour of functional coatings. In line with classic benchmarks, normally distributed noise is added with standard deviation of 0.5 and 2e5 for Ackley/Levy and SDL surfaces, respectively. Bounds for the x-dimensions of each surface can be found in the Appendix. We then compare the following search strategies:
\begin{enumerate}[(A)]
    \item \textbf{Random:} The worst-case baseline any strategy needs to beat. For each input parameter x, draw a sample from a uniform probability distribution with upper and lower bounds determined by the problem parameter space.
    \item \textbf{Expected Improvement (EI):} This is the classic EI implementation, using the analytic formula for candidate calculation.
    \item     \textbf{Noisy Expected improvement (qNEI):} This is the same as EI, but includes a model for the observed noise, as is commonly assumed in practical applications.
    \item \textbf{Mode cycling:} as proposed by \cite{rupnow2023self} whose data we use for our SDL test function. This method cycles between the upper confidence bound (UCB) acquisition function and a space-filling algorithm that selects a points furthest from all other points. 
\end{enumerate}

We choose the random search strategy as a worst-case baseline that every strategy should be able to beat. EI is the most popular search strategy \cite{frazier2018tutorial} and as such most relevant for comparison. qNEI is used to showcase a search strategy that accounts for noise, the standard setting SDLs operate in. The Mode-cycling is used to benchmark qNEI against a method that was used on a real-world SDL to handle asynchronous parallelisation. To counter the sub-optimal throughput of asynchronous parallel operation of a multi-stage SDL, we extend EI and qNEI with the implementation of a \textit{pending points masks} available in BoTorch \cite{balandat2020botorch}. This strategy modification prevents recommendation of candidates that are running but have pending results. We expect EI and qNEI extended with the pending points mask to yield the best performance across all strategies.
For each simulation, we average over 30 optimisation trials, using ten random initial points and a total of 100 observations. We look at delays $\in \{0,1,3,5,7\}$ and dimensions $\in \{3, 5, 7\}$, except for the SDL test function which is fixed at 7 dimensions. The length of the cycle in the Mode Cycle search strategy was varied depending on the amount of delay used in the simulation trial. For a delay of 0, 1, and 3, the mode cycle search strategy cycled between UCB with beta values 0.25, 2.5, and 25 and a space filling algorithm as described in \cite{rupnow2023self}. For a delay of 5, the mode cycle search strategy cycled between UCB with beta values 0.1, 0.25, 1, 2.5, and 25 and the space filling algorithm. For a delay of 7, the mode cycle search strategy cycled between UCB with beta values 0.1, 0.25, 0.5, 1, 2, 4, and 10 and the space filling algorithm. These choices are based on a heuristic approach to balance exploration and exploitation. For each simulation, a dataset of all x-values and their resulting y-value found from the test function are saved and analyzed.

\section{Results \& Discussion}

\begin{figure}[h]
\centering
\includegraphics[width=1\textwidth]{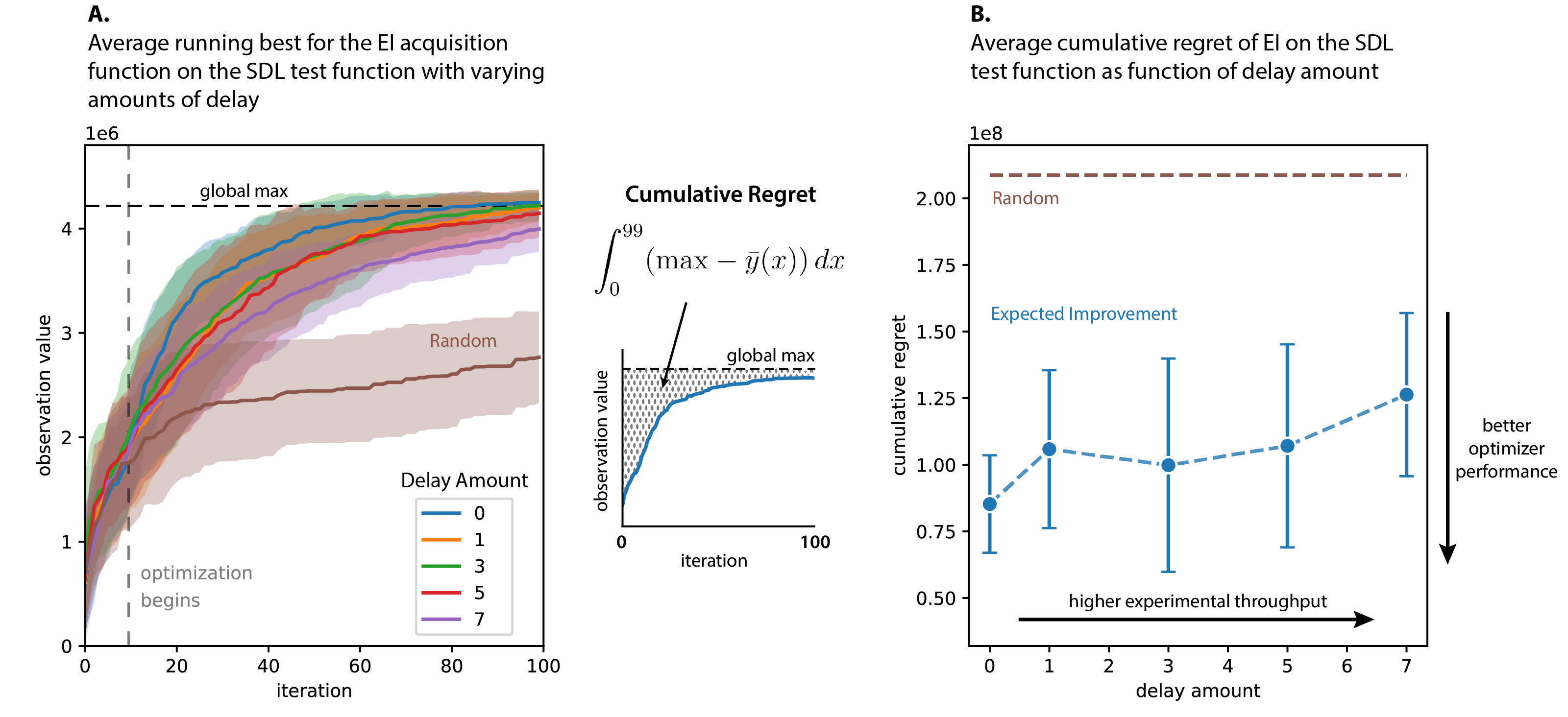}
\caption{(A) The EI acqusition function was used to optimise the SDL test function modeled using real SDL data. The average running best y-value is shown up to 100 iterations and averaged over 30 trials. The shaded region represents the standard deviation of the running best value at each iteration. (B) Cumulative regret was calculated by taking the area between the curve and the global maximum from observation 0 to 99 for each running best in (A). Cumulative regret is analogous to average optimiser performance, as a lower cumulative regret means that the optimiser has reached the global maximum in less iterations. The cumulative regret increases as the amount of delay increases. The average cumulative regret of the random baseline acquisition strategy is shown for comparison. The error bars represent the standard deviation of the cumulative regret over all thirty iterations.}
\label{fig:opt_delay}
\end{figure}

First, the performance loss of the simulated SDL due to delayed feedback was validated. The simulator was used to test the EI acquisition function on the SDL test function and the results are shown in figure \ref{fig:opt_delay}. The running best y-value is averaged over thirty trials and standard deviation is shown as the shaded region. It can be observed that optimisations with no delay find the global maximum in the fewest iterations on average. As the amount of delay increases, the number of iterations on average required to reach the global max increases. It should be noted that observations occasionally exceed the test function global maximum due to the addition of noise within the simulation. Exceeding a theoretical limit may be possible in real experiments as there may be measurement noise which may cause the observations to exceed the theoretical maximum.
The cumulative regret was then calculated and used as a metric of optimiser performance. The cumulative regret is calculated by calculating the area between the running best average curve and the global maximum from the first to the one hundredth observation. A lower cumulative regret is better because it means that the area is smaller and the optimiser was able to reach the global maximum faster. It can be seen in figure \ref{fig:opt_delay}B that the cumulative regret increases as the amount of delay increases. Because delay is associated with running experiments in asynchronous parallel, increased delay also typically means increased experimental throughput. This figure clearly illustrate the trade-off between experimental throughput and optimiser performance for an SDL operating experiments in asynchronous parallel. An SDL operator must balance this trade-off. For an expensive experiment, an SDL operator may choose to prioritize minimizing the number of experiments performed by running the SDL in serial operation as this will increase the likelihood to reach the global maximum in fewer experiments. For an inexpensive or exploratory experiment, it may be more practical to get as much data as possible and prioritize experimental throughput by increasing the amount of experiments running in parallel.

\begin{wrapfigure}{r}{0.5\textwidth}
  \centering
\includegraphics[width=0.42\textwidth]{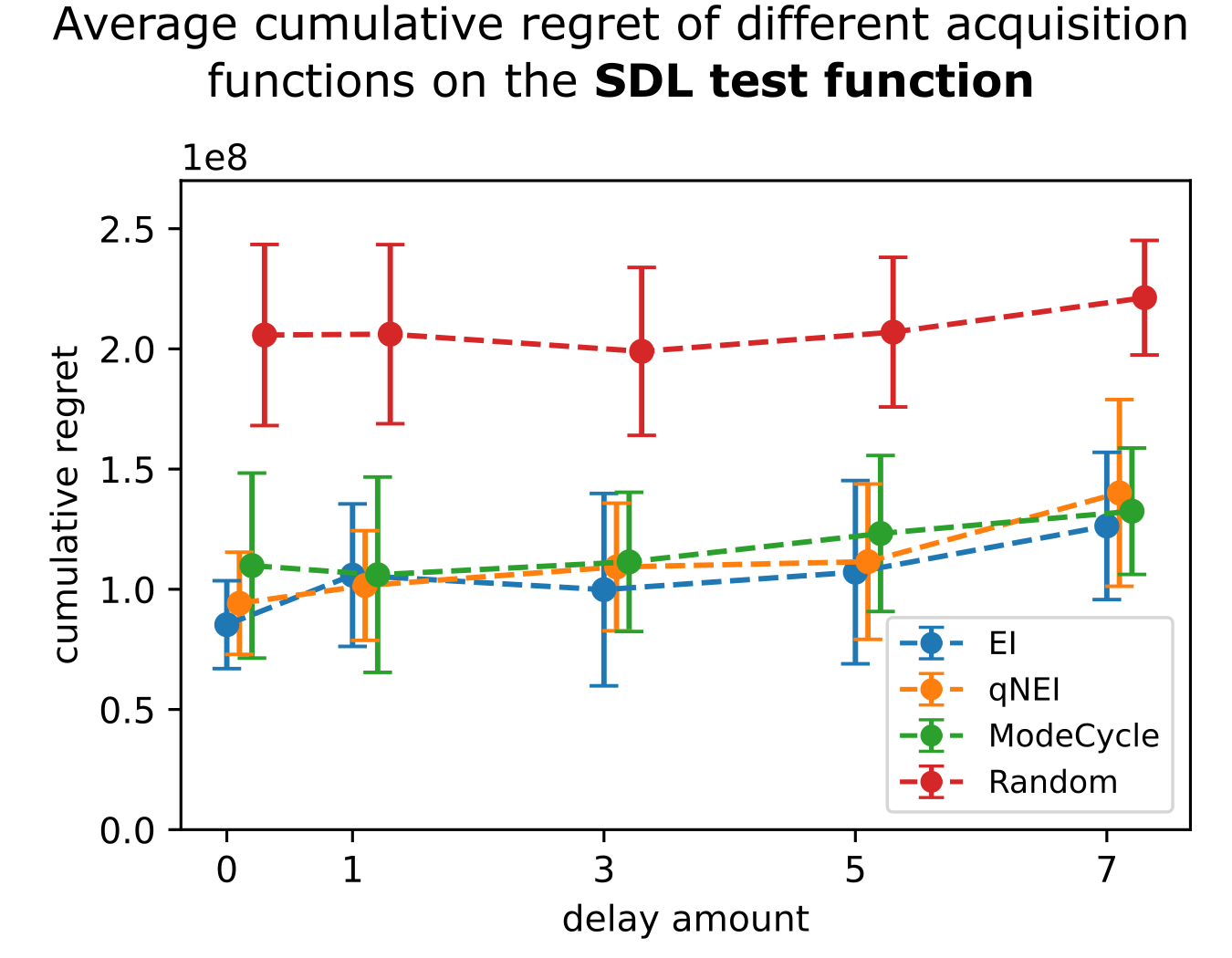}
\caption{Cumulative regret for different optimisation strategies on the SDL test function. The error bars represent the standard deviation of the cumulative regret over all thirty trials. The x-position of the data have been staggered to improve readability of the error bars.}
\label{SDL_cum}
\end{wrapfigure}

\begin{figure}[h]
\centering
\includegraphics[width=1\textwidth]{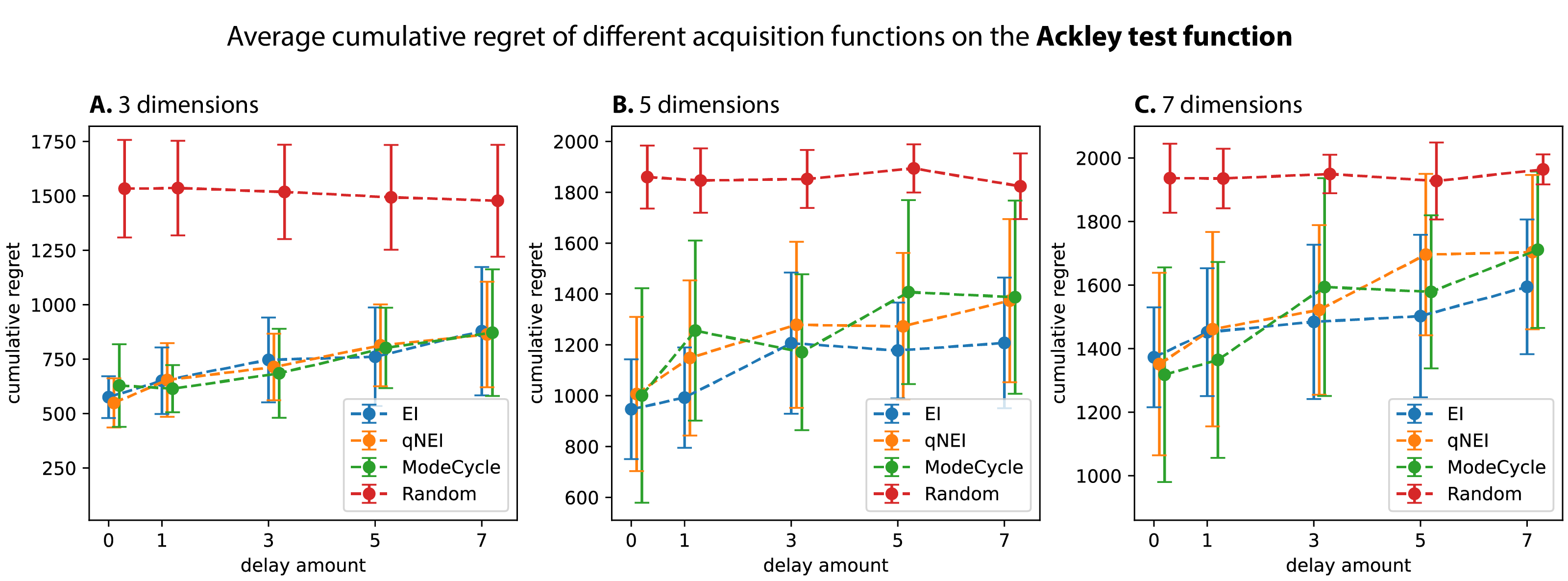}
\caption{Cumulative regret for different optimisation strategies on the Ackley test function in (A) 3, (B) 5, and (D) 7 dimensions. The error bars represent the standard deviation of the cumulative regret over all thirty trials. The x-position of the data have been staggered to improve readability of the error bars.}
\label{ackley_cum}
\end{figure}

Two additional acquisition functions were tested and compared: qNEI, to evaluate whether the noisy version could perform better, and ModeCycling, as proposed by Rupnow et al.\cite{rupnow2023self} as a way to reduce the repeated selection of identical experiments caused by delayed feedback. The cumulative regret of these tests are shown in \ref{SDL_cum}.  It can be seen that EI, qNEI, and ModeCycle all perform very similarly, with the cumulative regret increasing as delay amount increases. As expected, random sampling remains relatively constant since it does not depend on prior data to select new experiments. The average running best plots for each acquisition function and delay amount can be found in Appendix \ref{SDL_all}.
To validate that delay remains an issue on other optimisation problems, the same simulations were performed on the Ackley test function. Additionally, the number of dimensions was varied to get a sense of how dimensionality affects performance due to delay. The cumulative regret for each search strategy on the Ackley test function are shown in Figure \ref{ackley_cum}. It can be seen that as delay increases, the cumulative regret also increases linearly. As the dimensionality increases, the cumulative regret of EI, qNEI, and ModeCycle approach the cumulative regret of random sampling. The average running best plots for each dimension and delay amount for the Ackley function can be found in Appendix \ref{ackley_all}.

The cumulative regrets for each search strategy on the Levy test function are shown in Appendix Figure \ref{levy_cum}. The results are very similar to those of the Ackley function. However, increasing the number of dimensions in the Levy function significantly amplifies the cumulative regret. The average running best plots for each dimension and delay amount for the Levy function can be found in Appendix \ref{levy_all}.

Our simulation results showcase the trade-off between the efficiency gains of asynchronous parallel operation and performance loss due to delayed feedback. As expected, problems get harder with  increasing dimensionality. Additionally, as the number of parallel stages (i.e. delay) increases, performance also decreases. 
As expected, all search strategies outperform the random strategy in all cases. However, there is not enough evidence in our analysis to suggest that one acquisition strategy performs better than another in the presence of delay, especially in higher dimensions. To find an acquisition strategy that performs better in the presence of delayed feedback would require more trials to be run, an increased maximum number of iterations, and compared to a wider range of acquisition strategies. 
In the context of SDL operation, the decision whether to add delay to increase throughput or to add a dimension to increase the search area seems problem dependent – both increase the cumulative regret, though our tests found that adding delay seems to produce a smaller impact then adding dimensions. 

%Ideally, one can find a metric such as "adding one dimension to the problem is equivalent to adding 3 more stages". Of course, this is impossible with real-world SDLs, but choosing a test function close to the SDL problem space and running it with our code (made available after acceptance) should be a worthwhile investment to inform SDL design at least on an intuitive level. % I don't think this is necessary to say, particularly in the results section

\textbf{Limitations}
% limitations: Cum regret may not be the best for measuring optimizer performance. The optimizers have complicated behaviour that cannot be quanitified by a single number. Probability calculations should be performed to quantify the likelihood of hitting the max in a certain number of experiments.
The results presented here represent a simulated SDL and not a real SDL. However, we expect these results to hold true for a real SDL. It is important to remember, though, that the results presented here represent the average performance of each strategy.  An optimisation trial on a real SDL may perform better or worse than average. Additionally, cumulative regret is not the only metric for measuring optimiser performance. Optimisation algorithms have complicated behaviour that should not necessarily be quantified by a single number.
These results are true only for the three test functions presented here. Results may vary when exploring other material spaces or synthetic surfaces. Our work intends to guide SDL operators to use their machine more effectively and efficiently. Improvements in optimisation strategies will lead to better mathematical optimisations, but may or may not necessarily lead to higher performance materials found in fewer experiments. 

\textbf{Conclusion}
Considering the limited experimentation budgets commonly encountered in SDLs, simulations are an effective way to test search strategies in advance of deployment on an SDL. In this work we show that increasing throughput by running experiments in asynchronous parallel comes with a trade-off of reducing the performance of Bayesian optimisation algorithms. We  show that increasing delay by adding stages to an SDL produces a smaller impact to optimisation performance then adding dimensions to the materials problem space. We also compare a few different acquisition strategies, but were inconclusive in determining a champion strategy.
These results allow an SDL operator to make informed recommendations on which search strategy to use and — on average — find the global optimum faster than random guessing. As a next step, we are looking forward to collaborating on on implementing and evaluating these algorithms in real-world SDLs, including globally interconnected labs such as \cite{guevarra_kan_lai_jones_zhou_donnelly_richter_stein_gregoire_2023, strieth-kalthoff_hao_rathore_derasp_gaudin_angello_seifrid_trushina_guy_liu_et} which face similar asynchronous parallelization challenges, to further refine their search strategies and adjust them to specific applications. To facilitate this process, we are making a plug-in pipeline for the pending points implementation available online\footnote{\url{https://matterhorn.studio/pages/seminars/search-strategies-with-pending-points/}}. This will allow labs to easily access and integrate it into their SDLs.

\printbibliography

\section{Appendix}
%  Please include the references and supplementary materials in the same PDF as the main paper.

\textbf{Acknowledgements}
Thank you to the workshop reviewers for their detailed feedback that allowed us to improve and clarify our work. We are also grateful for the support for this work from the UKRI Innovate UK Transformative Technologies Grant 2023 Series.

\begin{figure}[h]
     \centering
     \begin{subfigure}[b]{0.49\textwidth}
         \centering
         \includegraphics[width=\textwidth]{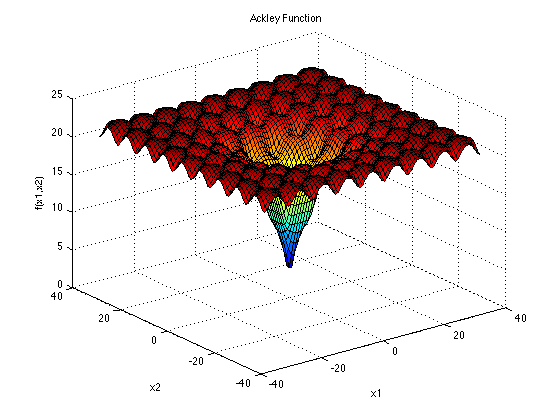}
         \caption{\textbf{Ackley}}
         \label{fig:ackley}
     \end{subfigure}
     \hfill
     \begin{subfigure}[b]{0.49\textwidth}
         \centering
         \includegraphics[width=\textwidth]{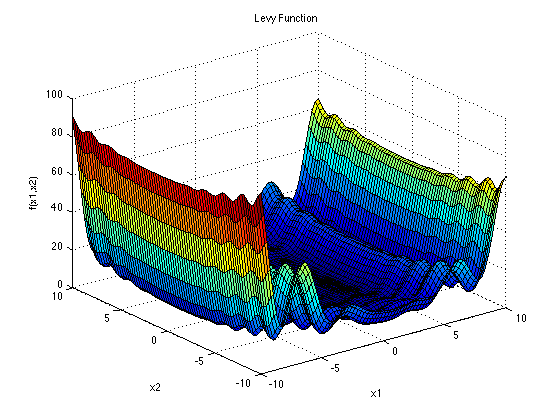}
         \caption{\textbf{Levy}}
         \label{fig:levy}
     \end{subfigure}
    \caption{Both Ackley (A, eq. \ref{eq:Ackley}) and Levy (B, eq. \ref{eq:Levy}) are comparatively challenging test functions that easily extend to higher dimensions \cite{test_functions}. Ackley is challenging for hill climbing algorithms that tend to get stuck in one of its many local minima, with its global minimum at $f(\mathbf{x}=0) = 0$. Levy is similar to Ackley but not symmetric around its global minimum at $f(\mathbf{x}=0) = 0$ and is often used to test algorithms aimed at handling high-dimensional problems with multiple local optima. In the context of our simulator, we chose to maximize the negative of each function.}
    \label{fig:test_functions}
\end{figure}

\textbf{Ackley test function}

\begin{equation}
    f(\mathbf{x}) = -a \cdot \exp \left(-b \sqrt{\frac{1}{d} \sum_{i=1}^{d} x_i^2}\right) - \exp \left(\frac{1}{d} \sum_{i=1}^{d} \cos(c x_i)\right) + a + exp(1)
    \label{eq:Ackley}
\end{equation}

where $a$, $b$, $c$ are constants that control the characteristics of the function and $d$ controls the dimensionality of the input space. We used the packaged BoTorch version of the function which defines $a=20$, $b=0.2$, and $c=2\pi$. The bounds for the Ackley test function are [-32.768, 32.768] for all $d$. The Ackley test function with $d=2$ can be seen in Figure \ref{fig:test_functions}A

\hfill \break
\textbf{Levy test function}

\begin{equation}
    f(\mathbf{x}) = \sin^2(\pi w_1) + \sum_{i=1}^{d-1} (w_i - 1)^2 \left[ 1 + 10 \sin^2(\pi w_i + 1) \right] + (w_d - 1)^2 \left[ 1 + \sin^2(2\pi w_d)\right]
    \label{eq:Levy}
\end{equation} 

where $w_i(x) = 1 + \frac{x_i - 1}{4}$, for all $i=1,...,d$. The bounds for the Levy test function are [-10, 10] for all $d$. The Levy test function with $d=2$ can be seen in Figure \ref{fig:test_functions}B

\textbf{SDL test function}

The SDL test function was based on a GP model built using training data from Rupnow et al. \cite{rupnow2023self} with a noise of 2e5 and fit using marginal log likelihood. The resulting model had cross validation $r^2$ of 0.9907. The bounds for the SDL test function shown below in Table \ref{tab:SDL_bounds}:
\begin{table}[h]
    \centering
    \caption{Bounds for the SDL test function}
    \begin{tabular}{c c c c c}
        {} & input variable & lower bound & upper bound & units \\
        \hline
        $x_1$ & DMSO content & 0 & 0.3 & v/v \\
        $x_2$ & precursor concentration & 10 & 20 & mg/mL \\
        $x_3$ & spray flow rate & 2 & 8 & µL/s\\
        $x_4$ & air flow rate & 65 & 100 & control valve \% \\
        $x_5$ & number of passes & 1 & 10 & passes \\
        $x_6$ & spray height & 10 & 25 & mm \\
        $x_7$ & hotplate temperature & 220 & 300 & °C \\
    \end{tabular}
    \label{tab:SDL_bounds}
\end{table}

\newpage
\textbf{Levy Cumulative Regret}

\begin{figure}[h]
\centering
\includegraphics[width=1\textwidth]{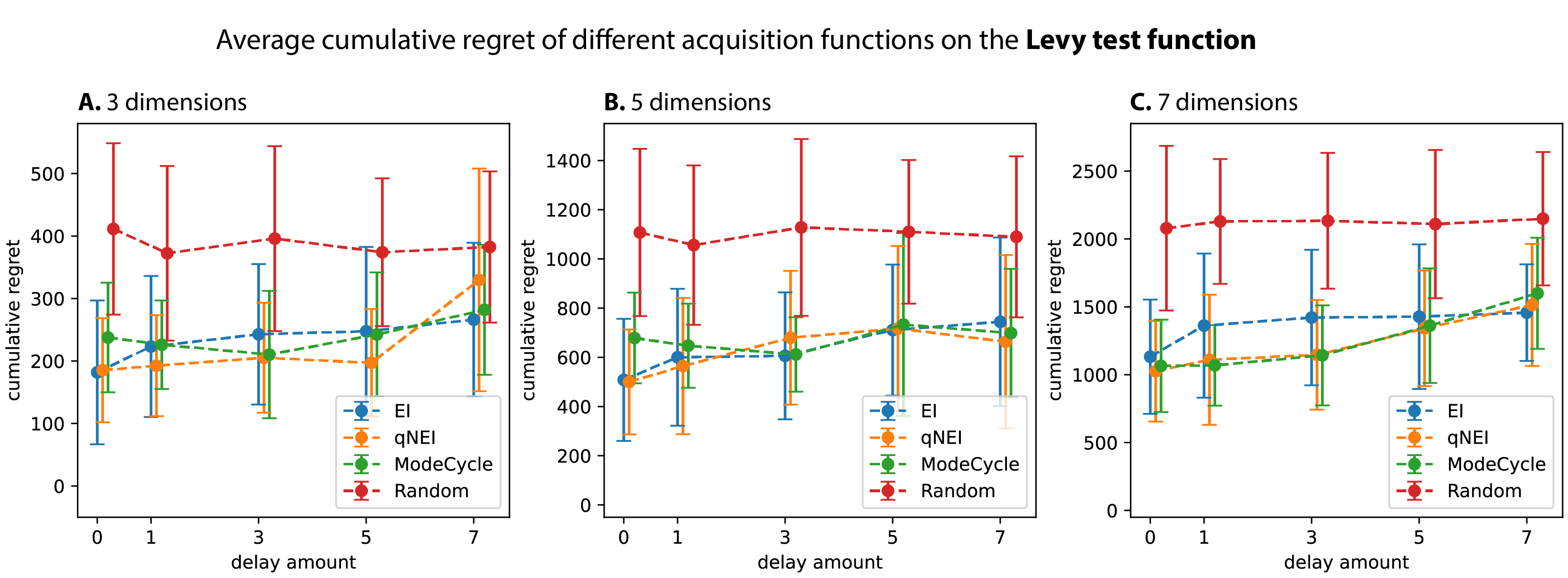}
\caption{Cumulative regret for different optimisation strategies on the Levy test function in (A) 3, (B) 5, and (D) 7 dimensions. The error bars represent the standard deviation of the cumulative regret over all thirty trials. The x-position of the data have been staggered to improve readability of the error bars.}
\label{levy_cum}
\end{figure}

\newpage
\textbf{Average running best optimisations}

\begin{figure}[h]
\centering
\includegraphics[width=1\textwidth]{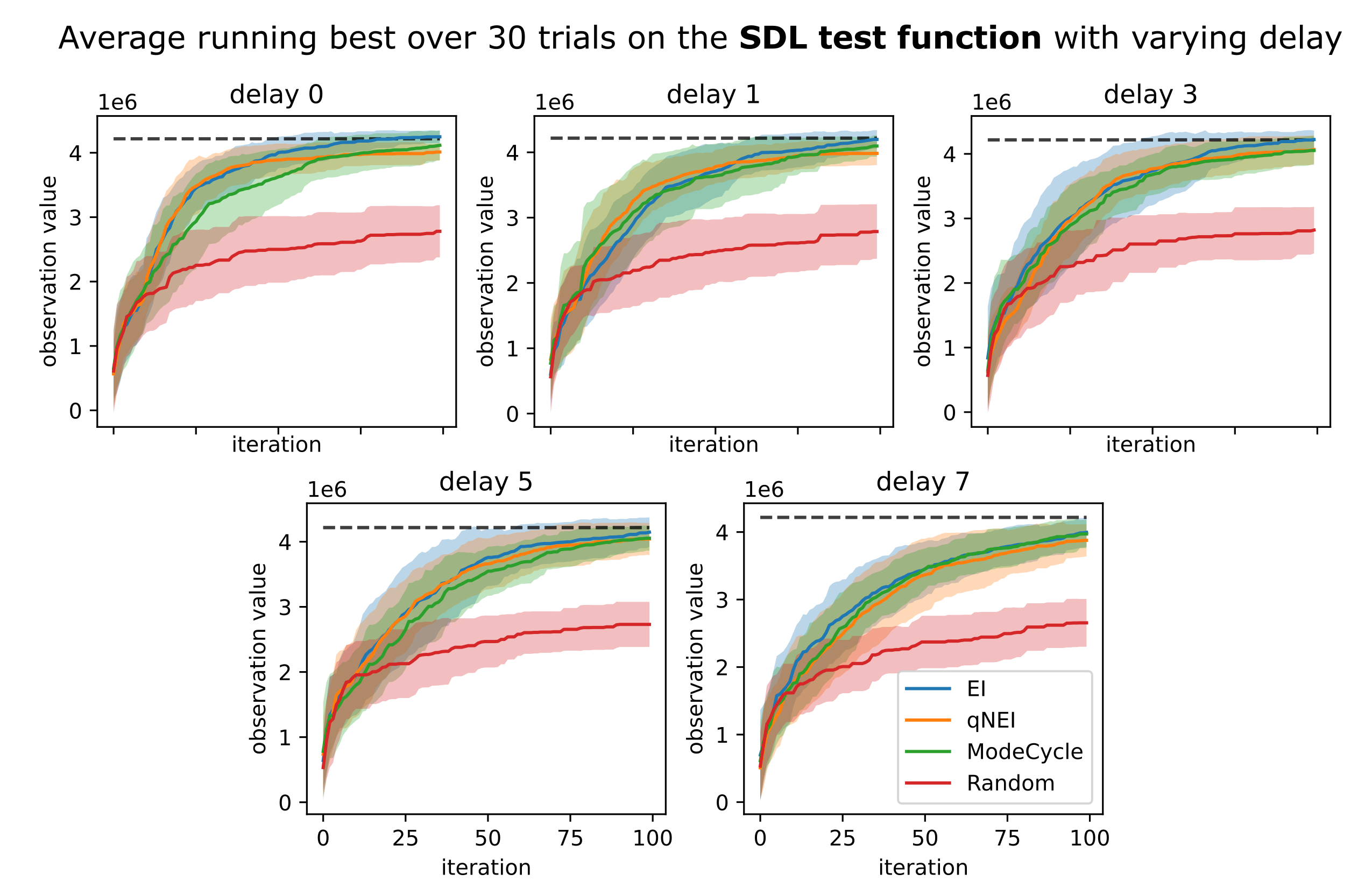}
\caption{Average running best observation over 30 optimisation trials on the SDL test function for delays $\in \{0,1,3,5, 7\}$. The line represents the mean running best observation while the shaded region represents the standard deviation. The dashed line represents the global maximum.}
\label{SDL_all}
\end{figure}

\begin{figure}[h]
\centering
\includegraphics[width=1\textwidth]{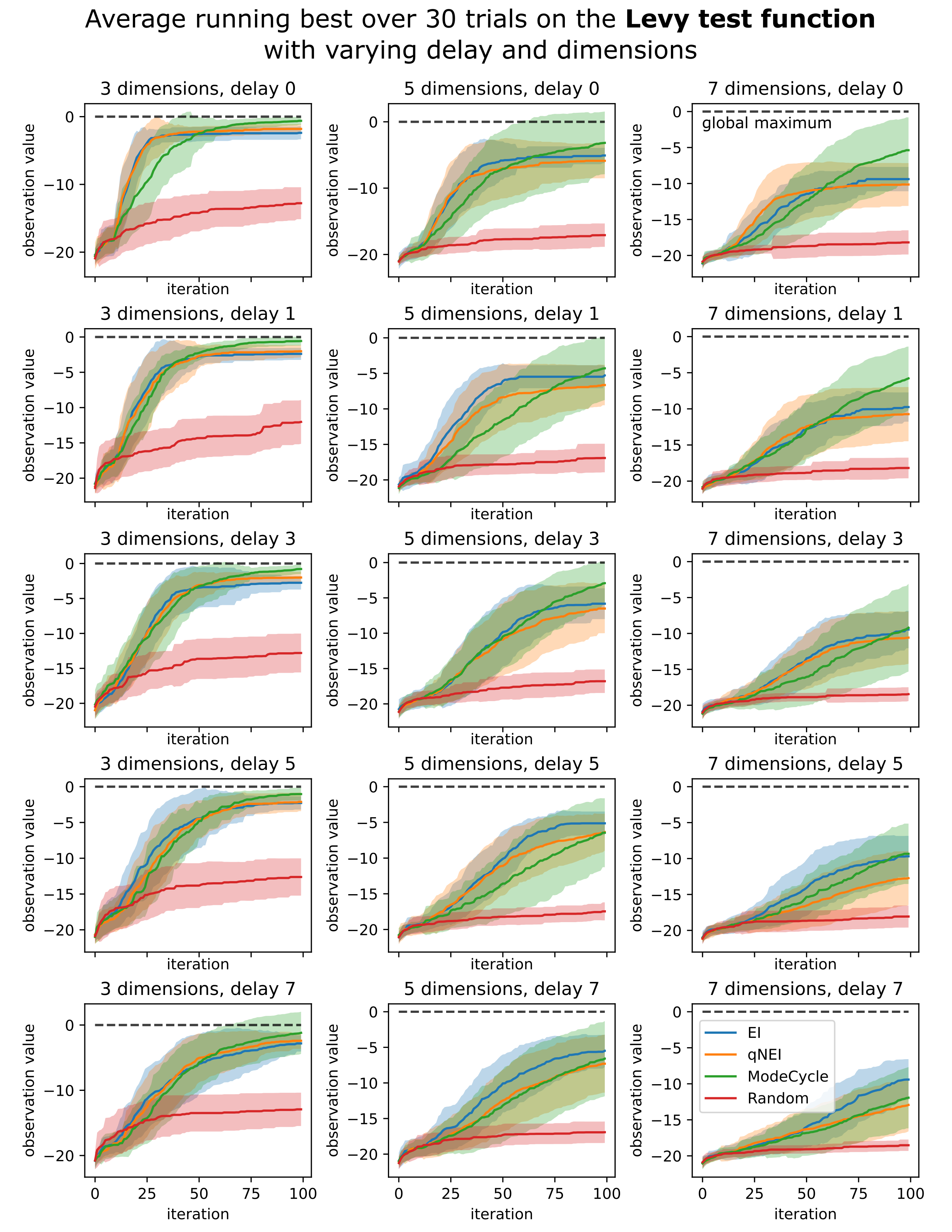}
\caption{Average running best observation over 30 optimisation trials on the Ackley test function for delays $\in \{0,1,3,5,7\}$ and dimensions $\in \{3,5,7\}$. The line represents the mean running best observation while the shaded region represents the standard deviation. The dashed line represents the global maximum.}
\label{ackley_all}
\end{figure}

\begin{figure}[h]
\centering
\includegraphics[width=1\textwidth]{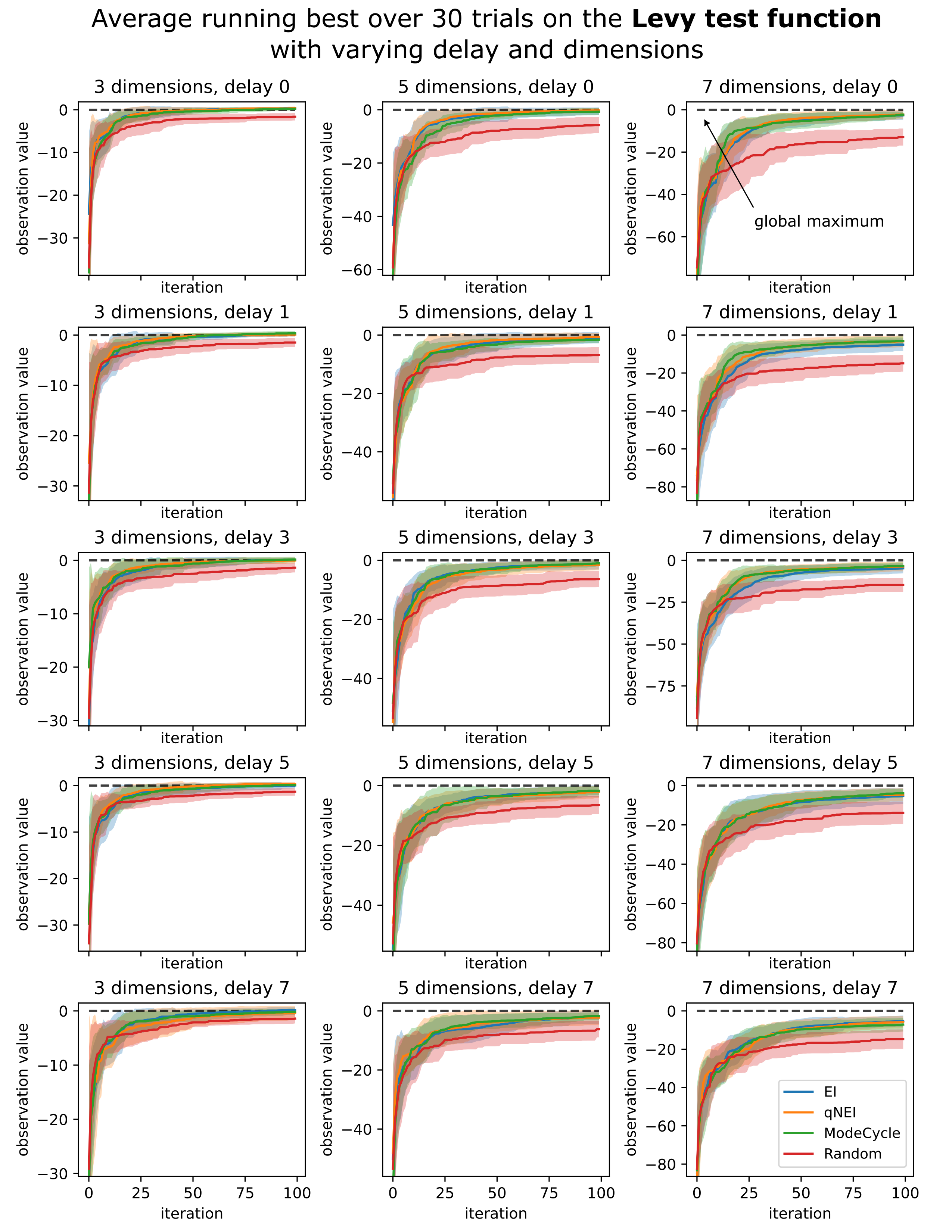}
\caption{Average running best observation over 30 optimisation trials on the Levy test function for delays $\in \{0,1,3,5,7\}$ and dimensions $\in \{3,5,7\}$. The line represents the mean running best observation while the shaded region represents the standard deviation. The dashed line represents the global maximum.}
\label{levy_all}
\end{figure}

\end{document}